%% file: paper.tex
\documentclass[conference]{IEEEtran}
\input{setup/packages}
\input{setup/macros}
\input{setup/acronyms}

\input{setup/title}

\begin{document}
\maketitle
\IEEEpubid{XXX-X-XXXX-XXXX-X/XX/\$NN.00 \copyright~2019 IEEE}

% %% %%%%%%%%%%%%%%%%%%%%%%%%%%%%%%%%%%%%%%%%%%%%%%%%%%%%
%
% Pages
%
% %% %%%%%%%%%%%%%%%%%%%%%%%%%%%%%%%%%%%%%%%%%%%%%%%%%%%%

\input{pages/abstract}
\input{pages/introduction}
\input{pages/document-structure}
\input{pages/athome}
\input{pages/hardware}
\input{pages/adopted-strategies}
\input{pages/challenges-internal}
\input{pages/conclusions}

% %% %%%%%%%%%%%%%%%%%%%%%%%%%%%%%%%%%%%%%%%%%%%%%%%%%%%%
%
% Final elements
%
% %% %%%%%%%%%%%%%%%%%%%%%%%%%%%%%%%%%%%%%%%%%%%%%%%%%%%%

\bibliographystyle{unsrtnatEtAl}
\footnotesize
\bibliography{paper}

\end{document}

%% file: setup/packages.tex
\usepackage{graphicx}
\usepackage[utf8]{inputenc}
\usepackage[greek,english,british,USenglish]{babel}
\usepackage{float}
\usepackage{ifthen}
\usepackage{xspace}
\usepackage{wrapfig}
\usepackage{xstring}
\usepackage{booktabs}
\usepackage{csquotes}
\usepackage{footnote}
\usepackage{hyphenat}
\usepackage{multirow}
\usepackage{ragged2e}
\usepackage{tabularx}
\usepackage{microtype}
\usepackage{marginnote}
\usepackage{subcaption}
\usepackage[T1]{fontenc}
\usepackage[all]{nowidow}
\usepackage{tablefootnote}
\usepackage[inline]{enumitem}
\usepackage[usenames,dvipsnames]{xcolor}
\usepackage[sort&compress,square,comma,numbers]{natbib}
\usepackage{pgfplots}
\usepackage{varioref}
\usepackage{hyperref}
\usepackage{cleveref}
\hypersetup{
	hidelinks=true,
	colorlinks=true,
	linkcolor={green!50!black},
	citecolor={blue!50!black},
	urlcolor={blue!80!black}
}

%% file: setup/macros.tex
\newcommand{\update}{%
}

\newcommand{\rpar}{)}% chktex 9

\newcommand\citereq[1]{%
	\ifthenelse{\equal{#1}{}}{%
		\textcolor{red}{[Citation Required!]}%
	}{%
		\textcolor{RedOrange}{[Need More Citations!]}%
		\cite{#1}%
	}%
}
\makeatletter
\@ifpackageloaded{natbib}{}{%
	\newcommand{\citep}[1]{\cite{#1}}%
	\newcommand{\Citep}[1]{\cite{#1}}%
	\newcommand{\citet}[1]{\cite{#1}}%
}
\newcommand\footnoteref[1]{\protected@xdef\@thefnmark{\ref{#1}}\@footnotemark}
\@ifundefined{url}{\newcommand\url[1]{#1}}{}
\makeatother

\newcommand{\formatmn}[1]{%
		\fontfamily{ptm}\selectfont%
		\bfseries\scriptsize#1%
}

\newcommand\athome{RoboCup@Home\xspace}

\makeatletter
\newcommand{\acronym}[3]{\newcommand{#1}[1][]{%
	\ifcsname @@#3\endcsname%
		\IfStrEqCase{##1}{%
			{n}{#3\marginnote{\formatmn{#3}}}%
			{f}{#2 (#3)}%
			{fn}{#2 (#3\marginnote{\formatmn{#3}})}%
			{x}{#2}%
			{xn}{#2\marginnote{\formatmn{#3}}}%
		}[#3]%
	\else%
		\IfSubStr{##1}{n}%
			{#2 (#3\marginnote{\formatmn{#3}})}%
			{#2 (#3)}%
		\expandafter\gdef\csname @@#3\endcsname{}%
	\fi\xspace%
}}
\makeatother

%% file: setup/acronyms.tex
\acronym{\wrs}{World Robot Summit}{WRS}
\acronym{\erl}{European Robotics League}{ERL}
\acronym{\tc}{Technical Committee}{TC}
\acronym{\oc}{Organizing Committee}{OC}
\acronym{\tdp}{Team Description Paper}{TDP}
\acronym{\tdps}{Team Description Papers}{TDPs}
\acronym{\mt}{Machine Translation}{MT}
\acronym{\hri}{Human-Robot Interaction}{HRI}
\acronym{\asr}{Automatic Speech Recognition}{ASR}
\acronym{\nlp}{Natural Language Processing}{NLP}
\acronym{\nlu}{Natural Language Understanding}{NLU}
\acronym{\ssl}{Sound Source Localization}{SSL}
\acronym{\spl}{Standard Platform Leagues}{SPL}
\acronym{\opl}{Open Platform League}{OPL}
\acronym{\dspl}{Domestic Standard Platform League}{DSPL}
\acronym{\sspl}{Social Standard Platform League}{SSPL}
\acronym{\gpsr}{\emph{General Purpose Service Robot}}{GPSR}
\acronym{\srad}{\emph{Speech Recognition and Audio Detection}}{SRAD}
\acronym{\egpsr}{\emph{Endurance General Purpose Service Robot}}{EGPSR}
\acronym{\eegpsr}{\emph{Enhanced Endurance General Purpose Service Robot}}{EEGPSR}

\acronym{\amcl}{Adaptive Monte Carlo Localization}{AMCL}
\acronym{\ann}{Artificial Neural Network}{ANN}
\acronym{\arnn}{Artificial Recursive Neural Network}{ARNN}
\acronym{\asp}{Answer Set Programming}{ASP}
\acronym{\brisk}{Binary Robust Invariant Scalable Keypoints}{BRISK}
\acronym{\brief}{Binary Robust Independent Elementary Features}{BRIEF}
\acronym{\clafu}{Classification Fusion}{CLAFU}
\acronym{\cnn}{Convolutional Neural Network}{CNN}
\acronym{\cuda}{Compute Unified Device Architecture}{CUDA}
\acronym{\dnn}{Deep Neural Network}{DNN}
\acronym{\dof}{Degrees of Freedom}{DoF}
\acronym{\dwa}{Dynamic Window Approach}{DWA}
\acronym{\fpfh}{Fast Point Feature Histograms}{FPFH}
\acronym{\hatp}{Hierarchical Agent-based Task Planner}{HATP}
\acronym{\hfsm}{Hierarchical Finite State Machine}{HFSM}
\acronym{\hog}{Histogram of Oriented Gradients}{HOG}
\acronym{\hrsi}{Human-Robot Spatial Interactions}{HRSI}
\acronym{\icp}{Iterative Closest Point}{ICP}
\acronym{\ism}{Implicit Shape Models}{ISM}
\acronym{\mrpt}{Mobile Robot Programming ToolKit}{MRPT}
\acronym{\ork}{Object Recognition Kitchen}{ORK}
\acronym{\pcl}{Point Cloud Library}{PCL}
\acronym{\ransac}{Random Sample Consensus}{RANSAC}
\acronym{\sift}{Scale-Invariant Feature Transform}{SIFT}
\acronym{\slam}{Simultaneous Localization and Mapping}{SLAM}
\acronym{\surf}{Speeded Up Robust Features}{SURF}
\acronym{\svm}{Support Vector Machine}{SVM}
\acronym{\tld}{Track-Learning Detection}{TLD}
\acronym{\ukf}{Unscented Kalman Filter}{UKF}
\acronym{\yolo}{You Only Look Once}{YOLO}
\acronym{\laser}{Laser Rangefinder Scanner}{LRFS}

%% file: setup/title.tex
\title{Trends, Challenges and Adopted Strategies in \athome%
	\\[-0.3cm]%
	{\normalsize --- 2019 arXiv version ---}
}

\author{
	\IEEEauthorblockN{
		Mauricio MATAMOROS\IEEEauthorrefmark{1},
		Viktor SEIB\IEEEauthorrefmark{1},
		and
		Dietrich PAULUS\IEEEauthorrefmark{1}
	}
	\IEEEauthorblockA{\IEEEauthorrefmark{1} Active Vision Group (AGAS), University of Koblenz-Landau. Universitätsstr. 1, 56070 Koblenz, Germany.}
}

%% file: pages/abstract.tex
% %% %%%%%%%%%%%%%%%%%%%%%%%%%%%%%%%%%%%%%%%%%%%%%%%%%%%%
%
% abstract.tex
%
% Author: Mauricio Matamoros
% Date:   2019.01.04
%
% %% %%%%%%%%%%%%%%%%%%%%%%%%%%%%%%%%%%%%%%%%%%%%%%%%%%%%
\begin{abstract}
Scientific competitions are crucial in the field of service robotics.
They foster knowledge exchange and benchmarking, allowing teams to test their research in unstandardized scenarios.
In this paper, we summarize the trending solutions and approaches used in \athome.
Further on, we discuss the attained achievements and challenges to overcome in relation with the progress required to fulfill the long-term goal of the league.
Consquently, we propose a set of milestones for upcoming competitions by considering the current capabilities of the robots and their limitations.

With this work we aim at laying the foundations towards the creation of roadmaps that can help to direct efforts in testing and benchmarking in robotics competitions.
\end{abstract}

%% file: pages/introduction.tex
% %% %%%%%%%%%%%%%%%%%%%%%%%%%%%%%%%%%%%%%%%%%%%%%%%%%%%%
%
% introduction.tex
%
% Author: Mauricio Matamoros
% Date:   2019.01.04
%
% %% %%%%%%%%%%%%%%%%%%%%%%%%%%%%%%%%%%%%%%%%%%%%%%%%%%%%
%chktex-file 1
\section{Introduction}
\label{sec:introduction}
Since its foundation in 2006, the \athome league has played an important role fostering knowledge exchange and research in service robotics.
Nowadays the competition influences ---and sometimes directs--- the course of research in the area of domestic service robotics.

Having such impact is not a minor thing.
The \athome league must take the responsibility of planning ahead and establish milestones for the competition.
This, of course, can only be done after analyzing the grounds in which the league is standing.

In response, we present a brief survey of the approaches and technical solutions reported by teams in each core functions required to accomplish a task, considering:
\begin{enumerate*}[label=\alph*\rpar]
	\item claims made in the \tdps,
	\item relevant publications,
	\item rulebooks,
	\item multimedia material available on-line,
	\item conducted polls targeting potential customers necessities,
	and
	\item our cumulative experience as participants and referees in RoboCup@Home since 2009.
\end{enumerate*}
Based on that survey we contribute by discuss the unconquered challenges that we have identified throughout these first twelve years.
Considering the studied material and the robot's current capabilities and achievements, elaborate on the next steps to take towards achieving the \athome goal, emphasizing those relevant in the short term.

%% file: pages/document-structure.tex
% %% %%%%%%%%%%%%%%%%%%%%%%%%%%%%%%%%%%%%%%%%%%%%%%%%%%%%
%
% document-structure.tex
%
% Author: Mauricio Matamoros
% Date:   2018.05.28
%
% %% %%%%%%%%%%%%%%%%%%%%%%%%%%%%%%%%%%%%%%%%%%%%%%%%%%%%
%chktex-file 1
This manuscript is organized as follows:
\Cref{sec:athome}, briefly introduces to \athome and its history.
\Cref{sec:hardware}, provides a brief summary of adopted hardware solutions.
\Cref{sec:strategies}, addresses the strategies and software solutions used by the participant teams.
In~\Cref{sec:challenges-intrinsic} we discuss the performance, challenges, and next steps.
Finally, in~\Cref{sec:conclusions} we summarize the discussion and present our final conclusions.

%% file: pages/athome.tex
% %% %%%%%%%%%%%%%%%%%%%%%%%%%%%%%%%%%%%%%%%%%%%%%%%%%%%%
%
% athome.tex
%
% Author: Mauricio Matamoros
% Date:   2018.05.28
%
% %% %%%%%%%%%%%%%%%%%%%%%%%%%%%%%%%%%%%%%%%%%%%%%%%%%%%%
%chktex-file 1
\section{\athome}
\label{sec:athome}
The \athome league was created in 2006 aiming to \enquote{develop service and assistive robot technology with high relevance for future personal domestic applications}
\footnote{\label{footnote:robocup-website}Source: \url{http://www.robocupathome.org/} Retrieved: Jan 1st, 2019.}.

The competition takes place in a testing arena shaped as a typical apartment.
In most tests, the robots solve household-related tasks, having their abilities and performance evaluated~\citep{Iocchi2015,Matamoros2018}.

During the first two years, the tests were scored with boolean criteria, making difficult to analyze the robot's performance.
Therefore, the scoring system was redesigned in 2008 and tests were split in a sequential set of goals while considering difficulty increases every third year~\citep{Wisspeintner2009,Iocchi2015,Matamoros2018}.

\begin{figure}[H]
	\centering
	\includegraphics[width=\linewidth]{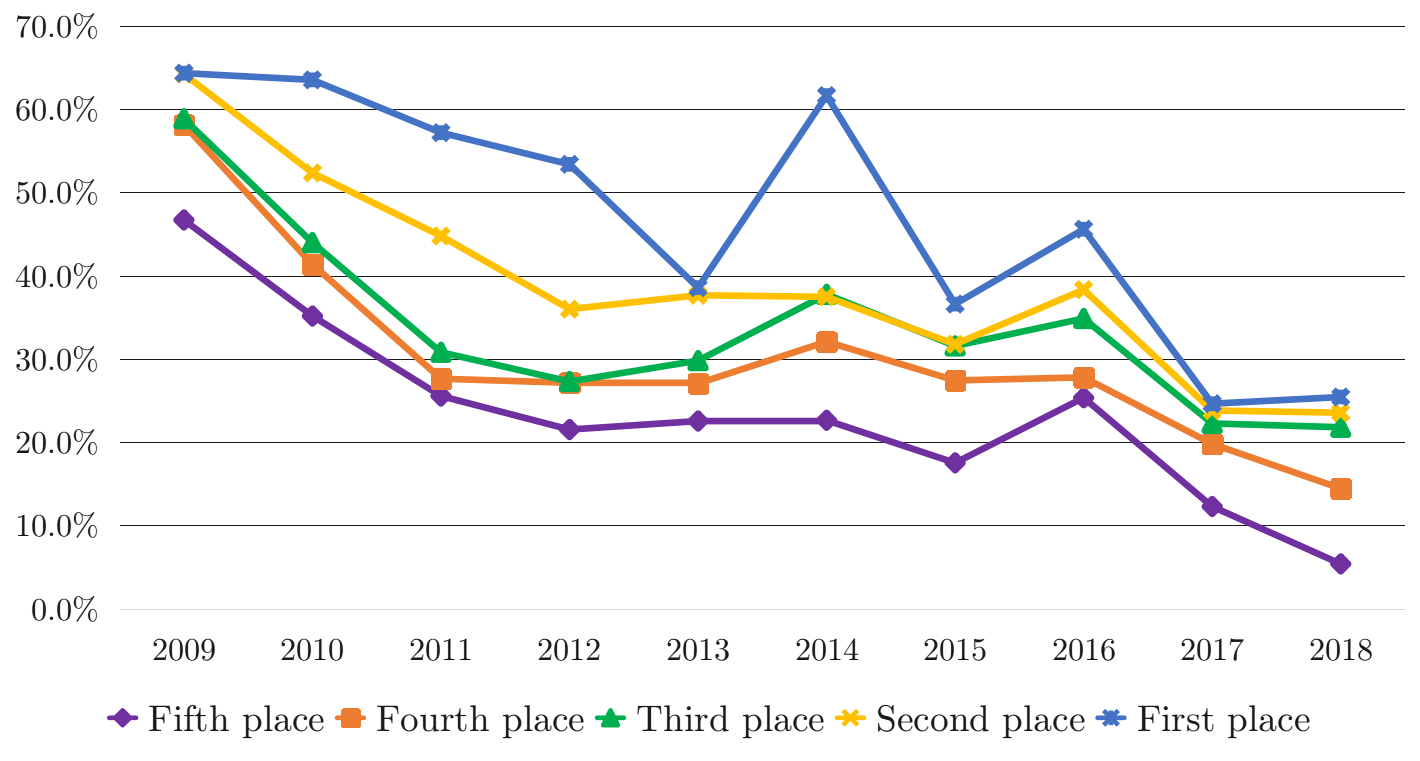}
	\caption{Performance of teams in the Top5 (2009--2018)}
	\label{fig:yearly-performance} %chktex 24
	{\centering\footnotesize Normalized values based on maximum attainable score}
\end{figure}

A sustained performance decrease (see~\Cref{fig:yearly-performance}) motivated the \tc to benchmark individual abilities.
Based on the results, in 2014 the test scheme proposed by~\citet{Wisspeintner2009} and analyzed by~\citet{Iocchi2015} was replaced by a new one focused in measuring performance~\citep{Matamoros2018}.

This new schema, considered ability benchmarking and repetitions to tackle the luck factor.
The introduced changes showed competitors their weaknesses, producing a performance increase in 2016 as~\Cref{fig:yearly-performance} shows, decreasing again in 2017 when difficulty was risen~\citep{Matamoros2018}.

Finally, in 2017 the league was split in three, the \opl with no hardware restrictions, the \dspl featuring the Toyota HSR, and the \sspl featuring the SoftBank-Robotics Pepper.
The division, veteran team migration, and the difficulty increase caused one of the lowest performances in the history of \athome~\citep{Matamoros2018}.

%% file: pages/hardware.tex
% %% %%%%%%%%%%%%%%%%%%%%%%%%%%%%%%%%%%%%%%%%%%%%%%%%%%%%
%
% hardware.tex
%
% Author: Mauricio Matamoros
% Date:   2018.06.07
%
% %% %%%%%%%%%%%%%%%%%%%%%%%%%%%%%%%%%%%%%%%%%%%%%%%%%%%%
%chktex-file 1
\section{Summary of Hardware Solutions}
\label{sec:hardware}

Here we summarize the hardware configurations most used in the \opl.
The aspects have been chosen for their potential influence in the robot's performance.

\begin{figure*}
	\begin{subfigure}[b]{0.32\textwidth}
		\centering
		\includegraphics[width=\textwidth]{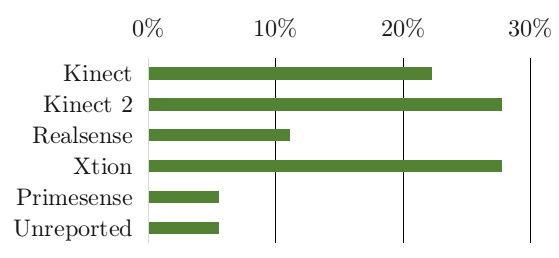}
		\caption{Reported RGB-D sensors}
		\label{fig:hardware-rgbd} %CHKTEX 24
	\end{subfigure}
	~ %chktex 39
	\begin{subfigure}[b]{0.32\textwidth}
		\centering
		\includegraphics[width=\columnwidth]{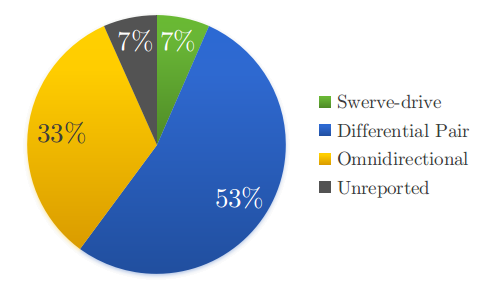}
		\caption{Reported locomotion types}
		\label{fig:hardware-base} %CHKTEX 24
	\end{subfigure}
	\begin{subfigure}[b]{0.32\textwidth}
		\centering
		\includegraphics[width=\columnwidth]{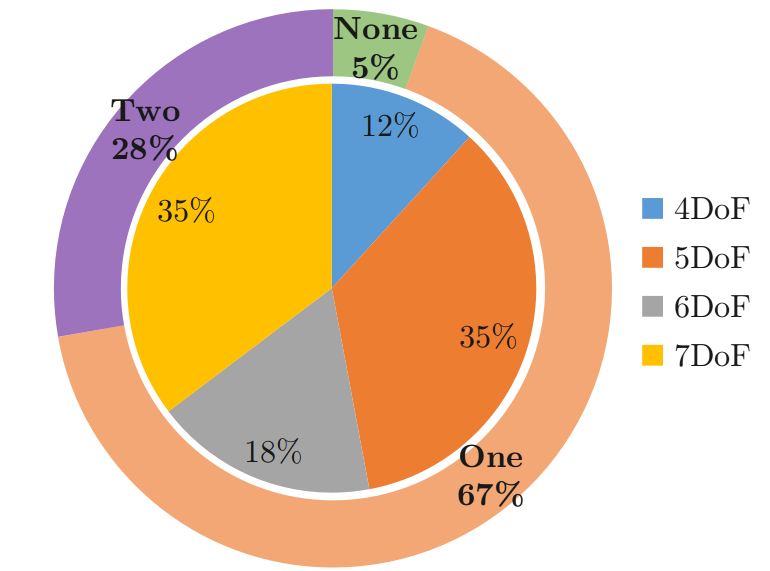}
		\caption{Reported solutions for manipulation}
		\label{fig:hardware-manipulator} %CHKTEX 24
	\end{subfigure}
	\caption{Reported hardware solutions (2017--2018)}
	\label{fig:hardware} %CHKTEX 24
\end{figure*}

\begin{enumerate}[nosep,leftmargin=1.5em,label=\arabic*.]
	\item \textbf{RGB-D Sensor:}
	\label{sec:hardware-rgbd} %CHKTEX 24
	In 2017 and 2018, all teams used at least one RGB-D sensor, preferring the Microsoft Kinect 2 due to its incorporated Time of Flight sensor and better resolution, with Asus Xtion in second place~(see~\Cref{fig:hardware-rgbd}).

	\item\textbf{Base:}%
	\label{sec:hardware-base} %CHKTEX 24
	All robots in \athome use wheels with no signs of upcoming changes (see~\Cref{fig:hardware-base}).
	The most used configuration is Differential Pair followed by Omni-Drive and with a single reported case of \textit{Swerve Drive}\footnote{Swerve drive is a special type of omni-directional configuration in which all four wheels rotate independently}.

	\item\textbf{Head:}%
	\label{sec:hardware-head} %CHKTEX 24
	To humans, head identification is intuitive, but robots can be deceiving. A robot can be \emph{bicephalous}
	or feature a face displaced from its sensors.
	Hence, we consider a robot head as a \emph{device integrated by one or more cameras and a pan-tilt unit with a microphone}.
	This criteria was met by 67\% (10 out 15) robots in Nagoya 2017 and 71\% (5 out 7) in Montreal 2018, all with 2 DoF.

	\item\textbf{Manipulator:}%
	\label{sec:hardware-manipulator} %CHKTEX 24
	To handle objects teams use either home-made or proprietary low-cost hardware~\Citep{tobi2017,uchile2017} since the size of professional arms often deems them unfit for domestic narrow spaces.\footnotemark
	Consequently, no professional manipulator was used in Nagoya 2017 nor in Montreal 2018.
	In both competitions the number of \dof[x] for manipulators ranged from 4 to 7~\Cref{fig:hardware-manipulator} depicts with an average strength in the final effector of $1.25kg$.
	Only one third of the robots featured two manipulators.
	\footnotetext{In 2016 team \emph{b-it-bots} experienced issues due to the size of the 7-DOF KUKA LWR  manipulator mounted on their Care-O-bot 3. The arm required a considerable amount of space to unfold and its size wouldn't make it fit in some spaces.}

	\item\textbf{Torso:}%
	\label{sec:hardware-torso} %CHKTEX 24
	We consider a robot torso as a \emph{device that provides panning and/or elevation to the robot's head and upper limbs}.
	In Nagoya 2017, only 60\% (9 out of 15) of competing robots had a torso, all cases featuring elevation only.
\end{enumerate}

%% file: pages/adopted-strategies.tex
% %% %%%%%%%%%%%%%%%%%%%%%%%%%%%%%%%%%%%%%%%%%%%%%%%%%%%%
%
% adopted-strategies.tex
%
% Author: Mauricio Matamoros
% Date:   2018.06.01
%
% %% %%%%%%%%%%%%%%%%%%%%%%%%%%%%%%%%%%%%%%%%%%%%%%%%%%%%
%chktex-file 1
\section{Adopted Strategies and Software Solutions}
\label{sec:strategies}
In this section, we summarize the solutions most commonly adopted by teams to address each of the basic functionalities involved in the tests.

\input{pages/stss-frameworks}
\input{pages/stss-navigation}
\input{pages/stss-people-recognition}
\input{pages/stss-object-recognition}
\input{pages/stss-asr-and-nlp}
\input{pages/stss-manipulation}

\begin{figure*}[t]
	\begin{subfigure}[t]{0.44\textwidth}
		\captionsetup{justification=centerfirst}%
		\caption{Navigation\\\footnotesize
		Report Ratio (R.R.) is number of  reporting \tdps}
		\label{tbl:strategies-nav} %CHKTEX 24
		\centering
		\small
		\begin{tabularx}{\columnwidth}{ X@{~} c l l }
			\toprule
			    \bfseries ~ &
				\bfseries R.R. &
				\bfseries Solution &
				\bfseries Cases \\
			\midrule
			Path planning      & 60\% & $A^*$          & (39\%) \\
			Obstacle avoidance & 55\% & ROS            & (42\%) \\
			~                  &      & Occupancy grid & (28\%) \\
			Localization       & 63\% & \amcl          & (58\%) \\
			Mapping            & 76\% & GMapping       & (69\%) \\
			\\
			\bottomrule
		\end{tabularx}
	\end{subfigure}
	~~ %chktex 39
	\begin{subfigure}[t]{0.25\textwidth}
		\captionsetup{justification=centering}%
		\caption{Object Recognition\\\footnotesize
		As reported by 84\% f the \tdps}
		\label{tbl:strategies-or} %CHKTEX 24
		\centering
		\small
		\begin{tabularx}{\columnwidth}{ X@{} c }
			\toprule
			\bfseries Solution & \bfseries Cases
			\\ \midrule
			\yolo        & 31\% \\
			\sift        & 19\% \\
			\surf        & 16\% \\
			OpenCV Caffe & 9\% \\
			Tensorflow   & 9\% \\
			\\
			\bottomrule
		\end{tabularx}
	\end{subfigure}
	~~ %chktex 39
	\begin{subfigure}[t]{0.26\textwidth}
		\captionsetup{justification=centering}%
		\caption{People Recognition\\\footnotesize
		As reported by 67\% of the \tdps}
		\label{tbl:strategies-pr}
		\centering
		\small
		\begin{tabularx}{\columnwidth}{ X@{} c }
			\toprule
			\bfseries Solution & \bfseries Cases
			\\ \midrule
			Openface              & 23\% \\
			Haar-based algorithms & 14\% \\
			Viola-Jones algorithm & 14\% \\
			Caffe                 &  9\% \\
			Microsoft Face        &  9\% \\
			OpenCV                &  9\% \\
			\bottomrule
		\end{tabularx}
	\end{subfigure}
	\caption{Adopted Software Solutions (2017--2018)}
	\label{tbl:strategies} %CHKTEX 24
\end{figure*}

%% file: pages/stss-frameworks.tex
% %% %%%%%%%%%%%%%%%%%%%%%%%%%%%%%%%%%%%%%%%%%%%%%%%%%%%%
%
% stss-architectures.tex
%
% Author: Mauricio Matamoros
% Date:   2018.06.07
%
% %% %%%%%%%%%%%%%%%%%%%%%%%%%%%%%%%%%%%%%%%%%%%%%%%%%%%%
%chktex-file 1
\subsection{Frameworks and Middlewares}
\label{sec:strategies-frameworks}
Nowadays, ROS has become a tacit standard in robotics with all teams using it at least partially~\Citep{Matamoros2018}.
Nonetheless, older frameworks like Orocos~\Citep{tue2018} are also used, while some teams continue using their own solutions~\Citep{pumas2018}, or vendor solutions like the NaoQi in \sspl.

%% file: pages/stss-navigation.tex
% %% %%%%%%%%%%%%%%%%%%%%%%%%%%%%%%%%%%%%%%%%%%%%%%%%%%%%
%
% stss-navigation.tex
%
% Author: Mauricio Matamoros
% Date:   2018.06.07
%
% %% %%%%%%%%%%%%%%%%%%%%%%%%%%%%%%%%%%%%%%%%%%%%%%%%%%%%
%chktex-file 1
\subsection{Navigation}
\label{sec:strategies-nav}
In \athome navigation involves
	path planning,
	obstacle avoidance,
	localization, and
	mapping.

Since safe, robust indoors navigation is taken for granted, path planning and localization are not scored inside the arena.
Contrastingly, obstacle avoidance is tested for small, reflecting, and hard-to-see object.
Finally, on-line mapping for unknown environments is being extensively tested.

With a couple of exceptions, it can be said that all teams rely on the ROS navigation stack (see~\Cref{tbl:strategies-nav}) tuned to the robot and based on each teams preferences~(see~\Cref{tbl:strategies-nav}).
Hence, the most broadly adopted solution sums up to OpenSlam's Gmapping and \amcl with an $A^*$ path planner~\Citep{tue2018,uchile2017,tobi2018}, with the incorporation of a Kalman filter to Gmapping as a good second~\Citep{tinker2017,walkingmachine2018}.

\subsubsection*{Path Planning and Obstacle Avoidance}
Several teams just report the use of ROS, so we assume an out-of-the-box configuration. %chktex 13
Solutions other than ROS include
Randomized Path Planners~\Citep{aisltut2017},
and
Wave-Propagation algorithms based on the Fast Marching Method~\Citep{northeastern2017}.
Regarding obstacle avoidance, most teams reported building occupancy grids based on the the \laser and the RGB-D camera.

\subsubsection*{Localization and Mapping}
These two abilities are often reported together.
Regarding localization alone, there are two reported solution differing from the robot's built-in localization and \amcl. These solutions are addressed by different teams and implement \slam using the \icp algorithm~\Citep{airobots2017,unsw2017}.
Both methods aim for accuracy and speed with limited resources.
On the mapping side, reported solutions other than GMapping include
hector \slam~\Citep{duckers2017}, \mrpt and \icp~\Citep{oittrial2017}, and Omnimapper~\Citep{tritonsdspl2017}.

%% file: pages/stss-people-recognition.tex
% %% %%%%%%%%%%%%%%%%%%%%%%%%%%%%%%%%%%%%%%%%%%%%%%%%%%%%
%
% stss-people-recognition.tex
%
% Author: Mauricio Matamoros
% Date:   2018.06.07
%
% %% %%%%%%%%%%%%%%%%%%%%%%%%%%%%%%%%%%%%%%%%%%%%%%%%%%%%
%chktex-file 1
\subsection{People Detection \& Recognition}
\label{sec:strategies-pr}
This section relates detection and recognition of static people using visual information.

Strategies for people detection combine face and skeleton detection to reduce false positives while limiting the detection range.
Further, there are hybrid techniques like combining 3D object recognition with face detection (e.g. OpenFace), or analysis of thermal images~\Citep{Iocchi2015,uchile2015,tue2018}.
Face detection strategies include
Openface~\Citep{tue2017},
Viola-Jones algorithm~\Citep{golem2017},
and
Haar-based algorithms~\Citep{happymini2018}.

In \athome, person recognition implies recalling the person's name, typically involving a facial recognition.
Notable solutions  include
\hog descriptors with \svm classifiers~\Citep{alle2017},
Strands Perception People~\Citep{tobi2018,spqrel2018},
Viola-Jones and Eigenfaces~\Citep{golem2017},
Siamese \cnn{}s~\Citep{aupair2018},
and
Haar Cascades with either EigenFaces~\Citep{pumas2018} or \dnn~\Citep{happymini2018}.
Some approaches consider texture and color segmentation as backup.
Recently, cloud services are being incorporated.

Finally, other features addressed in the competition consider height, gender, age, pose, relative position, and clothing; which are mostly tacled using \dnn{}-based libraries and cloud services~\Citep{hibikinodspl2017,homer2018}.

%% file: pages/stss-object-recognition.tex
% %% %%%%%%%%%%%%%%%%%%%%%%%%%%%%%%%%%%%%%%%%%%%%%%%%%%%%
%
% stss-object-recognition.tex
%
% Author: Mauricio Matamoros
% Date:   2018.06.07
%
% %% %%%%%%%%%%%%%%%%%%%%%%%%%%%%%%%%%%%%%%%%%%%%%%%%%%%%
%chktex-file 1
\subsection{Object Detection and Recognition}
\label{sec:strategies-or}
Although closely related, object detection is times faster than recognition, allowing continuous recognition.
Often, the point-cloud of the RGB-D sensor is used to remove background, floor, and other surfaces (e.g.~using Vector Quantization~\Citep{pumas2017} or \ransac~\Citep{pumas2018,northeastern2017}).
Then, color-depth images are clustered for its analysis by the object recognizer.
This approach is preferred over deep-learning-based for being less expensive.

Most teams use multiple algorithms either choosing by consensus or processing in a pipeline to add robustness to their object recognition.
Often depth information is used to recognize contours and shapes~\Citep{pumas2018,tritonsdspl2017,wrighteagle2017}.
Some solutions combine
\begin{enumerate*}[label=\alph*\rpar]
	\item \yolo with \sift and \brief~\Citep{aisltut2017},
	\item \surf with Continuous Hough-space voting with \ism~\Citep{homer2018},
	\item color, size and shape histograms and SIFT~\Citep{pumas2018},
	\item \pcl with \ransac and \yolo~\Citep{northeastern2017},
	and
	\item contours using LINEMOD with HSV color histograms and \surf considering joint models for occlusion~\Citep{wrighteagle2017}.
\end{enumerate*}

Although some teams run their algorithms on demand, more often processes are running all the time, although normally results are simply discarded by the task planner.
Very few teams have reported mechanisms to take advantage of continuous detection and recognition.

Beyond classification, object shape and feature recognition (e.g.~color, size, position, shape) is fundamental for grasping and describing previously untrained objects.
Unfortunately, the strategies used are not documented in the \tdps.

%% file: pages/stss-asr-and-nlp.tex
% %% %%%%%%%%%%%%%%%%%%%%%%%%%%%%%%%%%%%%%%%%%%%%%%%%%%%%
%
% stss-asr-and-nlp.tex
%
% Author: Mauricio Matamoros
% Date:   2018.06.07
%
% %% %%%%%%%%%%%%%%%%%%%%%%%%%%%%%%%%%%%%%%%%%%%%%%%%%%%%
%chktex-file 1
\subsection{Audio, Speech, and Natural Language Processing}
\label{sec:strategies-speech}
The most broadly adopted solution to deal with speech consists in a pipeline in which a filtered audio signal feeds an \asr[f] engine to get a text-transcript for further processing.
Then, the transcript is sent to a \nlp unit that extracts all relevant information that is passed to a high-level task planner that triggers the pertinent behaviors\cite{Matamoros2018rc,Kinarullathil2018}.

Often filtering is left to the microphone and the \asr engine~\cite{Doostdar2008}, being HARK\footnotemark a notable exception~\cite{aisltut2017,happymini2017,hibikinoOPL2017}.
\footnotetext{HARK (\textbf{H}onda Research Institute Japan \textbf{A}udition for \textbf{R}obots with \textbf{K}yoto University) is an open-source robot audition software that includes modules for \asr and sound-source localization and sound separation. Source: \url{https://www.hark.jp/}}

\begin{figure}
	\centering
	\includegraphics[width=\columnwidth]{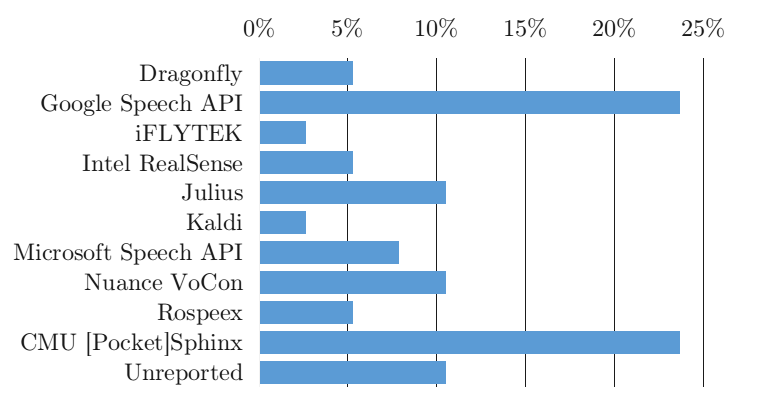}
	\caption{Trends in \asr[x]}
	\label{fig:asr-trends} %CHKTEX 24
\end{figure}

Regarding \asr, the most adopted off-line solutions include Julius~\cite{duckers2017}, the Microsoft Speech API~\cite{pumas2018}, and CMU Sphinx~\cite{kamerideropl2017,tobi2018,robofei2018} as~\Cref{fig:asr-trends} shows.
In \spl most teams use cloud services due to the limited computing power of standard robots. The Google speech API is the most popular approach~\cite{aisltut2017,jsk2017,aupair2018,spqrel2018,gentlebots2018}.
Due to network unreliability teams often have backup offline solutions.

Moving forward to \nlp[x], teams are migrating from keyword spotting and pattern matching with state machines~\cite{tinker2017,kamerideropl2017,Seib2015}.
Groups focusing in \nlp and \nlu are rare, and solutions are mentioned in less than 50\% of the \tdps.
Notable approaches include
Probabilistic Semantic Parsers~\cite{utaustinvilla2017},
Multimodal Residual \dnn~\cite{wrighteagle2017},
Ontology-based parsers with inference engines~\cite{pumas2017},
and
$\lambda$--calculus-based semantic parsing~\cite{aupair2017}.
The Stanford Parser~\cite{stanford2011corenlponline} is the most adopted solution for POS-tagging and syntactic tree extraction along with LU4R~\cite{Bastianelli2016}, a Spoken Language Understanding Chain for HRI~\cite{spqrel2018,walkingmachine2018}.

%% file: pages/stss-manipulation.tex
% %% %%%%%%%%%%%%%%%%%%%%%%%%%%%%%%%%%%%%%%%%%%%%%%%%%%%%
%
% stss-manipulation.tex
%
% Author: Mauricio Matamoros
% Date:   2018.06.07
%
% %% %%%%%%%%%%%%%%%%%%%%%%%%%%%%%%%%%%%%%%%%%%%%%%%%%%%%
%chktex-file 1
\subsection{Manipulation}
\label{sec:strategies-man}

A broadly adopted solution considers direct-inverse kinematic models with a closed-loop control and camera feedback as an alternative to the ROS manipulation stack, with many teams migrating to \textit{MoveIt!}

Manipulation was reported in only 18 (56\%) of the \tdps, \update
from which 61\% use \textit{MoveIt!} and the rest custom solutions.
These include
super-ellipsoid fitting~\Citep{tritonsdspl2017},
multiple deep auto-encoders fed with RGB-D and audio data~\Citep{erasers2017},
and
the proprietary built-in software~\Citep{spqrel2018}.

While two-handed manipulation is often attempted in the \opl, opening doors is still problematic.
In contrast, \dspl robots can open doors with ease\footnotemark~but lack of a second arm.
As of \sspl, manipulation is avoided due to the limitations of the robot.
\footnotetext{Doors with handles and lose latches as requires by the league and as shown on the HSR live demonstration for DSPL (\url{https://youtu.be/SwIkY1ffExI?t=30}) and in this other videos: \url{https://youtu.be/2nWX7-ccP8I?t=72} and \url{https://youtu.be/wCJ9qQrr5M0?t=46}.}

Remarkably, there is a growing trend for deep-learning-based methods, especially in planning.
Although computationally more expensive, they can be faster and more robust than the traditional methods when a non-optimal solution is acceptable.
Besides, although their supporters claim they are easier to develop, the required data acquisition for training is their biggest weakness~\citep{Kim2018}.

%% file: pages/challenges-internal.tex
% %% %%%%%%%%%%%%%%%%%%%%%%%%%%%%%%%%%%%%%%%%%%%%%%%%%%%%
%
% challenges-internal.tex
%
% Author: Mauricio Matamoros
% Date:   2018.06.07
%
% %% %%%%%%%%%%%%%%%%%%%%%%%%%%%%%%%%%%%%%%%%%%%%%%%%%%%%
%chktex-file 1
\section{Discussion on Intrinsic Challenges}
\label{sec:challenges-intrinsic}
The challenges described in this section correspond to the role of the \athome in directing research in robotics.
Here, we discuss the most relevant abilities.

\input{pages/cint-navigation}
\input{pages/cint-people-recognition}
\input{pages/cint-object-recognition}
\input{pages/cint-manipulation}
\input{pages/cint-asr-and-nlp}
\input{pages/cint-roadmaps}

%% file: pages/cint-navigation.tex
% %% %%%%%%%%%%%%%%%%%%%%%%%%%%%%%%%%%%%%%%%%%%%%%%%%%%%%
%
% cint-navigation.tex
%
% Author: Mauricio Matamoros
% Date:   2018.06.07
%
% %% %%%%%%%%%%%%%%%%%%%%%%%%%%%%%%%%%%%%%%%%%%%%%%%%%%%%
%chktex-file 1
\subsection{Navigation}
\label{sec:challenges-nav}
Although closely related, indoor and outdoor navigation are approached with disjoint ability sets.

\subsubsection*{Indoor-Navigation}
\label{sec:challenge-nav-in}
Despite the continuous improvements, steps, wet floors, rough carpets, and in general uneven surfaces present restrictions that haven't been addressed yet.
Furthermore, potential customers prefer silent carpet-friendly robots, specially when considering night cleaning.

Also troublesome are the amount of geometrical data required and the sensor's perception range cap of 4m that might impede localization in wide spaces (e.g.~big houses and facilities).
Hence semantic localization must be incorporated, specially to move around regular buildings.
We found reading signs and labels is also an important feature in offices and hospitals.

Finally important abilities discontinued or not addressed yet include using an elevator, navigate in narrow spaces, move furniture around, and functional touching (push aside objects with one's body).
People often rely on their body to move or stop objects when moving around.

\subsubsection*{Outdoor-Navigation}
\label{sec:challenge-nav-out}
Even in urban environments, exteriors add an extra tier of complexity.
Deal with rough terrain (e.g.~grass, gravel, and sand), slopes, far distances, sunlight, and weather conditions is necessary to meet customers expectations.
For instance, pet owners are interested in robots capable of walking and exercise the animals.

As distances and the number of stimuli to process grows, so does the computational power required.
Robots need to handle occlusions and and react to fast-moving objects in real time with higher precision than autonomous cars, while at the same time consider signs, read street and shop names, and prevent collisions with children, elders, dogs, other robots, etc.
Furthermore, the robot needs to synchronize with streetlights, automatic doors, and public transport.

\subsubsection*{Conclusions}
\label{sec:challenge-nav-conclusions}
Although \athome nowadays features only wheeled-robots, this design is not suitable for most human environments.
Consequently either environments must become robot-friendly, or robots will need other mechanisms (i.e.~legs).

Finally, robots need to be programmed to recognize and map semantic information in their surroundings like people does, correlating perception with previous knowledge.

%% file: pages/cint-people-recognition.tex
% %% %%%%%%%%%%%%%%%%%%%%%%%%%%%%%%%%%%%%%%%%%%%%%%%%%%%%
%
% cint-people-recognition.tex
%
% Author: Mauricio Matamoros
% Date:   2018.06.07
%
% %% %%%%%%%%%%%%%%%%%%%%%%%%%%%%%%%%%%%%%%%%%%%%%%%%%%%%
%chktex-file 1
\subsection{People Detection \& Recognition}
\label{sec:challenges-pr}
Command retrieval, object delivery, and \hri make people detection a must-have function.
As of 2018, robots must detect still people either standing, sitting or lying; exhibiting robust, faulty, and bad performance on each pose respectively.

In contrast, people recognition is barely addressed.
In this analysis we split the feature in two aspects, people recalling and identification.
\textit{Recalling} usually consists in pairing the person's features (typically the facial ones) with other information, such as a name or an order.
Key problems strive in the reduced number of people (less than five) and the briefness of the data lifespan (information is disposed after a test).

Relatively new, \textit{people identification} relates to finding a person matching a description, or describing individuals or groups of people.
In 2018, provided descriptions weren't accurate
and description-based people search was tested at random.
In consequence, it is impossible to give precise information regarding general performance.
Nonetheless, it must be acknowledged that most robots perform good at counting groups of people and identifying their gender.

\subsubsection*{Next Steps}
We believe the league should emphasize people recognition.
Features like estimated age, gender, and relative position can be used to test awareness, decision making and planning, as well as integrating voice recognition (recalling).
Furthermore, as with objects, detection ranges need to be increased, as well as the amount of people to remember.

We also consider important to introduce the detection of features like: emotions, moods, activities, attire styles, clothing elements, healthiness, and vital signs (inebriation, fatigue, sickness, sleep, etc).
Also important, and shared with object recognition is addressing occlusions and translucent surfaces.

\subsubsection*{Conclusion}
It's clear that tests require to be redesigned to ensure benchmarking data is available for analysis.
In addition, we find that visual and audio information are equally important when addressing people.

%% file: pages/cint-object-recognition.tex
% %% %%%%%%%%%%%%%%%%%%%%%%%%%%%%%%%%%%%%%%%%%%%%%%%%%%%%
%
% cint-object-recognition.tex
%
% Author: Mauricio Matamoros
% Date:   2018.06.07
%
% %% %%%%%%%%%%%%%%%%%%%%%%%%%%%%%%%%%%%%%%%%%%%%%%%%%%%%
%chktex-file 1
\subsection{Object Detection and Recognition}
\label{sec:challenges-or}
As of 2018, it is safe to state that robots are shortsighted.
Despite the remarkable advances, detection is often constrained to 4 meters due to RGB-D sensor resolution and reach, causing robots to fail at \emph{seeing} objects lying in direct line of sight.
Further objects are always placed sparsely, being partially occluded at worst, and a robot must know about 25 of them, a situation that greatly differs from most people's homes.

Remarkably, object detection is being used to aid in navigation by updating the grid of an occupancy map.
In addition, robots are starting to successfully identify of objects \textit{of a kind} (e.g.~apples, regardless the color, size etc).

Notably, all reported strategies feed-forward process the input stream frame-by-frame, discarding used visual information.

\subsubsection*{Next Steps}
\label{sec:challenges-or-next-steps}
Brain inspired models from neurosciences suggest the use of contextual information to build scene representations, so processes can rely on memory instead of visual search~\Citep{Oliva2004,Oliva2007}, leading to a faster response.
In addition, processed visual information could be used for reinforcement learning and update the world model instead of being discarded.

Other problems to address include
dust, spots, and dirt detection,
recognize stacked objects,
position-based target selection among equals,
increase the number of objects,
and recognize objects in odd lightning conditions like direct sunlight and in the dark.
Similarly, translucent and transparent objects present new challenges when it comes to occlusions and detection of spots.

Likewise, and borderline with manipulation, relevant problems include
orientation identification,
best placement location identification, and
weight inference.

\subsubsection*{Conclusions}
\label{sec:challenges-or-conslusions}
Like with other abilities, a roadmap is necessary to direct advances on object recognition.
The first logical steps seems to be enhance the aspects that help with manipulation and navigation like placing surfaces, grasping orientation, and obstacle identification; as well as occlusions.

Finally, it is necessary to start integrating object recognition with high-level action planning so semantic, spatial and temporal relationships can be also recognized.
A far-placed milestone would include objects that change over time (e.g.~food).

%% file: pages/cint-manipulation.tex
% %% %%%%%%%%%%%%%%%%%%%%%%%%%%%%%%%%%%%%%%%%%%%%%%%%%%%%
%
% cint-manipulation.tex
%
% Author: Mauricio Matamoros
% Date:   2018.06.07
%
% %% %%%%%%%%%%%%%%%%%%%%%%%%%%%%%%%%%%%%%%%%%%%%%%%%%%%%
%chktex-file 1
\subsection{Manipulation}
\label{sec:challenges-man}
Manipulation is the most mature research area in robotics,
nonetheless \athome is far from reaching the feats of industrial robots due to the lack of a fixed inertial base and a concealed working space.

As of 2018 most manipulation tasks are special cases of \emph{fetch} and \emph{pick-and-place}, although robots in \dspl and \opl are required to open the door of a cupboard, a subtask that was skipped by all \opl participants in 2017 and by most of them in 2018~\Citep{Matamoros2018}.
Also skipped were pouring, spot scrubbing, and tray transport~\Citep{Matamoros2018}.
Nonetheless, although skipped in 2017, several teams did it remarkably well at handling cutlery in 2018~\Citep{Matamoros2018}.
Furthermore the competition considers mostly moving regular-shaped, small (5--25cm, 2lt max volume), non-fragile lightweight objects (75--950gr) like apples, small cereal boxes, and [empty] soda cans.

\subsubsection*{Market Requirements}
Our polls reveal people are explicitly asking for robots that can
clean the toilet, wipe windows, do the dishes (by hand), wash, iron and fold clothes, open flasks and jars, brush and wash the dog, and take out the garbage to name some examples.

Other important skills required in daily use include opening doors, move furniture, and operate the controls of electrical appliances.
From these, opening doors has been addressed since 2006, but not really solved until team eR@sers impressed the league in 2016 with the proprietary robot now used in the \dspl[x].
Notwithstanding, it seems there is still a long way to go before this skill can be considered solved in \opl and \sspl.

\subsubsection*{Next Steps}
The manipulation capabilities of the robots must be expanded in several directions follows:
\begin{itemize}[nosep,leftmargin=1.5em]
	\item \textbf{Reach:} As of 2018, objects are placed between $30cm$ and $1.8m$ height and no further than $5cm$ from the border.
	Next steps involve reaching the floor and up to $50cm$ in depth.

	\item \textbf{Maneuverability:} Nowadays robot's movement is restricted to flat surfaces and hand-overs.
	Hence picking from/placing into boxes and bags is required.
	At the same time, twisting, uncapping, shaking, folding, levering, and turning are features yet to be tested in \athome.

	\item \textbf{Strength:} Terminal-effector's load must be gradually increased to reach at least $3kg$ when fully extended, and beyond.
	Heavy loads are required by elders for grocery transport and help when stand up.

	\item \textbf{Precision:}
	Torque, speed, trajectory, and acceleration control are required to spread butter, whip cream, and grasp pills.
	Besides, some people requested application of insulin injections.
\end{itemize}

Finally, other important tasks include
manipulating a switch,
taking out garbage,
serve soup,
mopping,
unpacking, and
passing towels.

\subsubsection*{Conclusion}
The \athome league would enormously benefit from a manipulation roadmap with which teams could plan hardware improvements with sufficient time.
For now, gradually increasing weight and placing distance would suffice.

%% file: pages/cint-asr-and-nlp.tex
% %% %%%%%%%%%%%%%%%%%%%%%%%%%%%%%%%%%%%%%%%%%%%%%%%%%%%%
%
% cint-asr-and-nlp.tex
%
% Author: Mauricio Matamoros
% Date:   2018.06.07
%
% %% %%%%%%%%%%%%%%%%%%%%%%%%%%%%%%%%%%%%%%%%%%%%%%%%%%%%
%chktex-file 1
\subsection{Speech and Natural-Language Interaction}
\label{sec:challenges-asr}
A key aspect to people's communication, speech in \athome is mostly restricted to issuing commands to robots~\cite{Matamoros2018,Matamoros2018rc}.
These commands are generated using grammars given in advance to teams under noise conditions which are far from those found in domestic environments~\cite{Matamoros2018,Matamoros2018rc}.
Further, experts take part in all interactions, meaning that the way to address a robot is not entirely natural~\cite{Matamoros2018,Matamoros2018rc}.

An thorough study, available in~\cite{Matamoros2018rc}, was conducted addressing these troubling aspects in spoken \hri, followed by a roadmap and a set of strategies to tackle them and foster dialog-based natural language interactions in \athome.

It's worth mention, nonetheless, that so far available \asr engines are an important limiting factor in natural language interaction.

%% file: pages/cint-roadmaps.tex
% %% %%%%%%%%%%%%%%%%%%%%%%%%%%%%%%%%%%%%%%%%%%%%%%%%%%%%
%
% cint-roadmaps.tex
%
% Author: Mauricio Matamoros
% Date:   2018.06.07
%
% %% %%%%%%%%%%%%%%%%%%%%%%%%%%%%%%%%%%%%%%%%%%%%%%%%%%%%
%chktex-file 1
\subsection{Roadmaps and milestones}
\label{sec:challenges-roadmaps}
We believe the \athome must take the responsibility of leading and guide research in domestic service robotics.
This implies carefully planning milestones and scheduling tests for the competition.
Said in other works, the league should have roadmaps to direct research and help during test design.

To teams, knowing the challenges in advance will help them to prepare direct their research, and invest better their resources.
As for \tc an official roadmap can prevent newly elected members from overriding previous decisions, damping the league's progress.

In consequence, appearing valuable ideas need to be analyzed, evaluated against the robot's capabilities, and condensed in milestones and test drafts to be retaken in latter years.

As response, in this paper we have condensed teams' ideas and concerns, the user's needs, and the current capabilities and limitations of the robots; presenting them in~\Cref{sec:challenges-pr,sec:challenges-or,sec:challenges-nav,sec:challenges-man} as future steps or milestones .
Nonetheless, this is just a first step, since roadmaps need to be still prepared and adopted by the league which, in the end, is peer-maintained.

%% file: pages/conclusions.tex
% %% %%%%%%%%%%%%%%%%%%%%%%%%%%%%%%%%%%%%%%%%%%%%%%%%%%%%
%
% conclusions.tex
%
% Author: Mauricio Matamoros
% Date:   2018.05.28
%
% %% %%%%%%%%%%%%%%%%%%%%%%%%%%%%%%%%%%%%%%%%%%%%%%%%%%%%
%chktex-file 1
\section{Discussion and Conclusions}
\label{sec:conclusions}
In this paper we conduct a thorough summary of the software solutions and strategies used by participating teams in \athome to address some of the most important abilities required in the competition.
Further on, we present an overview of the attained achievements since the league's foundation based on our experience as long-time participants, contributors, and referees in the league.
Finally, also organized per skill and along with the overview, we discuss these achievements while addressing what is expected by potential consumers, what needs to be done, and what would be the next logical steps based on the robot's current capabilities.

This study result in two main contributions.
First, we believe the presented summary can serve as quick reference guide for new competitors, or for experienced ones looking for alternatives to their current solutions.
Second, our work sets the basis to build roadmaps to plan the direction and impact of robotics competitions like \athome.
Moreover, roadmaps can direct scientific research and help teams to prepare years in advance by establishing mid- and log-term goals, and perhaps a smarter resource management.

However, there is still work to be done.
Not only the roadmaps have to be designed.
This work has also allowed us to identify several important flaws that need to be addressed.
For instance, the presence of certain rules might be undermining the development of certain features.
At the same time, we have found that many successful approaches and strategies are never reported in the \tdps.
This has two important setbacks.
First, it makes it much harder for the scientific community to compare the performance of the different approaches when the best performers are missing.
Second, it leads to an eventual loss of knowledge that worsens as the lifetime of a good team shortens.

These insights are left to the competition organizers to analyze as part of future work, for which we trust this manuscript can come handy.

%% file: paper.bbl
\begin{thebibliography}{46}
\providecommand{\natexlab}[1]{#1}
\providecommand{\url}[1]{\texttt{#1}}
\expandafter\ifx\csname urlstyle\endcsname\relax
  \providecommand{\doi}[1]{doi: #1}\else
  \providecommand{\doi}{doi: \begingroup \urlstyle{rm}\Url}\fi

\bibitem[Iocchi et~al.(2015)Iocchi, Holz, Ruiz-del Solar, Sugiura, and Van
  Der~Zant]{Iocchi2015}
Luca Iocchi, et~al.
\newblock Robocup@home: Analysis and results of evolving competitions for
  domestic and service robots.
\newblock \emph{Artificial Intelligence}, 229:\penalty0 258--281, 2015.

\bibitem[Matamoros et~al.(2018{\natexlab{a}})Matamoros, Seib, Memmesheimer, and
  Paulus]{Matamoros2018}
Mauricio Matamoros, et~al.
\newblock Robocup@home: Summarizing achievements in over eleven years of
  competition.
\newblock In \emph{2018 IEEE International Conference on Autonomous Robot
  Systems and Competitions (ICARSC)}, pages 186--191, April 2018{\natexlab{a}}.
\newblock \doi{10.1109/ICARSC.2018.8374181}.

\bibitem[Wisspeintner et~al.(2009)Wisspeintner, Van Der~Zant, Iocchi, and
  Schiffer]{Wisspeintner2009}
Thomas Wisspeintner, et~al.
\newblock Robocup@home: Scientific competition and benchmarking for domestic
  service robots.
\newblock \emph{Interaction Studies}, 10\penalty0 (3):\penalty0 392--426, 2009.

\bibitem[Wachsmuth et~al.(2017)Wachsmuth, Lier, Meyer~zu Borgsen, Kummert,
  Lach, and Sixt]{tobi2017}
Sven Wachsmuth, et~al.
\newblock Tobi-team of bielefeld: The human-robot interaction system for
  robocup @home 2017.
\newblock 2017.

\bibitem[Mart{\'i}nez et~al.(2017)Mart{\'i}nez, Mu{\~n}oz, Olave, Pais, Hernan,
  Gomez, Garrido, Campanini, Orellana, Loncomilla, and Ruiz-del
  Solar]{uchile2017}
Luz Mart{\'i}nez, et~al.
\newblock Uchile homebreakers 2017 team description paper.
\newblock \emph{RoboCup @Home 2017 Team Description Papers}, 2017.

\bibitem[{van der Burgh} et~al.(2018){van der Burgh}, Lunenburg, Appeldoorn,
  Wijnands, Clephas, Baeten, van Beek, Ottervanger, Aleksandrov, Assman, Dang,
  Geijsberts, Janssen, {van Rooy}, Hofkamp, and {van de Molengraft}]{tue2018}
M.F.B. {van der Burgh}, et~al.
\newblock Tech united eindhoven @home 2018 team description paper.
\newblock \emph{RoboCup@Home 2018 Team Description Papers}, 2018.

\bibitem[Savage et~al.(2018)Savage, Martell, Estrada, Negrete, Cruz, Cruz,
  Cruz, Vazquez, Marquez, Silva, Pano, and Alvarez]{pumas2018}
Jesus Savage, et~al.
\newblock Pumas@home 2018 team description paper.
\newblock \emph{RoboCup@Home 2018 Team Description Papers}, 2018.

\bibitem[Wachsmuth et~al.(2018)Wachsmuth, Lier, and Meyer~zu Borgsen]{tobi2018}
Sven Wachsmuth, Florian Lier, and Sebastian Meyer~zu Borgsen.
\newblock Tobi - team of bielefeld: A human-robot interaction system for
  robocup@home 2018.
\newblock \emph{RoboCup@Home 2018 Team Description Papers}, 2018.

\bibitem[Guo et~al.(2017)Guo, Yao, Ma, Dong, Zhu, Peng, Wang, and
  Ma]{tinker2017}
Jiacheng Guo, et~al.
\newblock Tinker@home 2017 team description paper.
\newblock \emph{RoboCup @Home 2017 Team Description Papers}, 2017.

\bibitem[Cousineau and La~Madeleine(2018)]{walkingmachine2018}
Jeffrey Cousineau and Philippe La~Madeleine.
\newblock Walking machine @home 2018 team description paper.
\newblock \emph{RoboCup@Home 2018 Team Description Papers}, 2018.

\bibitem[Oishi et~al.(2017)Oishi, Miura, Koide, Demura, Kohari, Une,
  Villamar-Gomez, Kato, Kojima, and Morohashi]{aisltut2017}
Shuji Oishi, et~al.
\newblock Aisl-tut @home league 2017 team description paper.
\newblock \emph{RoboCup @Home 2017 Team Description Papers}, 2017.

\bibitem[Kelestemur et~al.(2017)Kelestemur, Allaban, and
  Padir]{northeastern2017}
Tarik Kelestemur, Anas~Abou Allaban, and Taskin Padir.
\newblock Frasier: Fostering resilient aging with self-efficacy and
  independence enabling robot team northeastern’s approach for robocup@home.
\newblock \emph{RoboCup @Home 2017 Team Description Papers}, 2017.

\bibitem[Li et~al.(2017)Li, Liu, Su, Li, Cheng, Hsieh, Liang, Chen, Lin, and
  Chang]{airobots2017}
Tzuu-Hseng~S. Li, et~al.
\newblock airobots\_ncku 2017 team description paper.
\newblock \emph{RoboCup @Home 2017 Team Description Papers}, 2017.

\bibitem[Sammut et~al.(2017)Sammut, Pagnucco, Castro, Flannagan, Gratton,
  Hengst, Rajaratnam, Schwering, Thielscher, Velonaki, and Wiley]{unsw2017}
C.~Sammut, et~al.
\newblock Unsw robocup@home spl team description paper.
\newblock \emph{RoboCup @Home 2017 Team Description Papers}, 2017.

\bibitem[Isobe et~al.(2017)Isobe, Katsumata, Tabuchi, Kinose, Ishimine, Fukui,
  Kobayashi, Taniguchi, Aly, Hagiwara, and Taniguchi]{duckers2017}
Shota Isobe, et~al.
\newblock Duckers 2017 @home dspl team description paper.
\newblock \emph{RoboCup @Home 2017 Team Description Papers}, 2017.

\bibitem[Miyawaki et~al.(2017)Miyawaki, Sano, Inoue, Hiroi, Nishiguchi, and
  Suzuki]{oittrial2017}
Kenzaburo Miyawaki, et~al.
\newblock O.i.t-trial 2017 team description paper.
\newblock \emph{RoboCup @Home 2017 Team Description Papers}, 2017.

\bibitem[Christensen et~al.(2017)Christensen, Riek, White, Parashar, Wang,
  Iqbal, Taylor, and Chan]{tritonsdspl2017}
Henrik~I. Christensen, et~al.
\newblock Uc san diego 2017 team description paper.
\newblock \emph{RoboCup @Home 2017 Team Description Papers}, 2017.

\bibitem[Mart{\i}nez et~al.(2015)Mart{\i}nez, Pavez, Olave, Correa,
  S{\'a}nchez, Loncomilla, and Ruiz-del Solar]{uchile2015}
Luz Mart{\i}nez, et~al.
\newblock Uchile homebreakers 2015 team description paper.
\newblock 2015.

\bibitem[van~der Burgh et~al.(2017)van~der Burgh, Lunenburg, Appeldoorn,
  Wijnands, Clephas, Baeten, van Beek, Ottervanger, van Rooy, and van~de
  Molengraft]{tue2017}
MFB van~der Burgh, et~al.
\newblock Tech united eindhoven @home 2017 team description paper.
\newblock \emph{University of Technology Eindhoven}, 2017.

\bibitem[Pineda et~al.(2017)Pineda, Rasc{\'o}n, Fuentes, Rodr{\'i}guez, Ortega,
  Reyes, Hern{\'a}ndez, Cruz, V{\'e}lez, and Ram{\'i}rez]{golem2017}
Luis~A. Pineda, et~al.
\newblock The golem team, robocup@home 2017.
\newblock \emph{RoboCup @Home 2017 Team Description Papers}, 2017.

\bibitem[Demura et~al.(2018)Demura, Enomoto, Nagashima, Yamakawa, Iwasaki, and
  Mashimo]{happymini2018}
Kosei Demura, et~al.
\newblock Kit happy robot 2018 team description.
\newblock \emph{RoboCup@Home 2018 Team Description Papers}, 2018.

\bibitem[Cheng et~al.(2017)Cheng, Ramirez-Amaro, Dianov, Lanillos, Guadarrama,
  Dean, Lozinska, Diez-Valencia, Grzywok, Guo, Simonic, and Wang]{alle2017}
Gordon Cheng, et~al.
\newblock Alle@home 2017 team description paper.
\newblock \emph{RoboCup @Home 2017 Team Description Papers}, 2017.

\bibitem[Lázaro et~al.(2018)Lázaro, Iocchi, Nardi, Fentanes, and
  Hanheide]{spqrel2018}
M.T. Lázaro, et~al.
\newblock Spqrel 2018 team description paper.
\newblock \emph{RoboCup@Home 2018 Team Description Papers}, 2018.

\bibitem[Choi et~al.(2018)Choi, Lee, Lee, Kim, and Zhang]{aupair2018}
Jinyoung Choi, et~al.
\newblock Team aupair technical paper for robocup@home 2018 social standard
  platform league.
\newblock \emph{RoboCup@Home 2018 Team Description Papers}, 2018.

\bibitem[Hori et~al.(2017{\natexlab{a}})Hori, Ishida, Kiyama, Tanaka, Kuroda,
  Hisano, Imamura, Himaki, Yoshimoto, Aratani, Hashimoto, Iwamoto, Morie, and
  Tamukoh]{hibikinodspl2017}
Sansei Hori, et~al.
\newblock Hibikino-musashi@home spl 2017 team description paper.
\newblock \emph{RoboCup @Home 2017 Team Description Papers},
  2017{\natexlab{a}}.

\bibitem[Memmesheimer et~al.(2018)Memmesheimer, Wettengel, Debald, Eckert,
  Möhlenhof, Evers, Heuer, Theisen, Buchhold, Eisenmenger, Häring, and
  Paulus]{homer2018}
Raphael Memmesheimer, et~al.
\newblock Robocup 2018 - homer@unikoblenz (germany).
\newblock \emph{RoboCup@Home 2018 Team Description Papers}, 2018.

\bibitem[Savage et~al.(2017)Savage, Negrete, Cruz, Marquez, Martell, Cruz,
  Vazquez, Pano, Cruz, Silva, Estrada, Arce, Matamoros, Garzon, and
  Fuentes]{pumas2017}
Jesus Savage, et~al.
\newblock Pumas@home 2017 team description paper.
\newblock \emph{RoboCup @Home 2017 Team Description Papers}, 2017.

\bibitem[Liu et~al.(2017)Liu, Zhang, Tang, and Chen]{wrighteagle2017}
Jiangchuan Liu, et~al.
\newblock Wrighteagle@home 2017 team description paper.
\newblock \emph{RoboCup @Home 2017 Team Description Papers}, 2017.

\bibitem[Matamoros et~al.(2018{\natexlab{b}})Matamoros, Harbusch, and
  Paulus]{Matamoros2018rc}
Mauricio Matamoros, Karin Harbusch, and Dietrich Paulus.
\newblock From commands to goal-based dialogs: A roadmap to achieve natural
  language interaction in robocup@home.
\newblock 2018{\natexlab{b}}.

\bibitem[Kinarullathil et~al.(2018 (in press))Kinarullathil, Martins, Azevedo,
  Lima, Lawless, Lima, Custódio, and Ventur]{Kinarullathil2018}
Mithun Kinarullathil, et~al.
\newblock From user spoken commands to robot task plans: a case study in
  robocup@home.
\newblock In \emph{IROS Workshop on Language and Robotics}, 2018 (in press).

\bibitem[Doostdar et~al.(2009)Doostdar, Schiffer, and Lakemeyer]{Doostdar2008}
Masrur Doostdar, Stefan Schiffer, and Gerhard Lakemeyer.
\newblock A robust speech recognition system for service-robotics applications.
\newblock In Luca Iocchi, et~al., editors, \emph{RoboCup 2008: Robot Soccer
  World Cup XII}, pages 1--12, Berlin, Heidelberg, 2009. Springer Berlin
  Heidelberg.
\newblock ISBN 978-3-642-02921-9.

\bibitem[Demura et~al.(2017)Demura, Demura, Nagashima, Enomoto, Yamakawa,
  Iwasaki, and Mashimo]{happymini2017}
Kosei Demura, et~al.
\newblock Happy mini 2017 team description paper.
\newblock \emph{RoboCup @Home 2017 Team Description Papers}, 2017.

\bibitem[Hori et~al.(2017{\natexlab{b}})Hori, Ishida, Kiyama, Tanaka, Kuroda,
  Hisano, Imamura, Himaki, Yoshimoto, Aratani, Hashimoto, Iwamoto, Morie, and
  Tamukoh]{hibikinoOPL2017}
Sansei Hori, et~al.
\newblock Hibikino-musashi@home 2017 team description paper.
\newblock \emph{RoboCup @Home 2017 Team Description Papers},
  2017{\natexlab{b}}.

\bibitem[Tan et~al.(2017)Tan, Duan, Ismail, and Uchimura]{kamerideropl2017}
Jeffrey Too~Chuan Tan, et~al.
\newblock Kamerider opl @home 2017 team description paper.
\newblock \emph{RoboCup @Home 2017 Team Description Papers}, 2017.

\bibitem[Gonbata et~al.(2018)Gonbata, Neves, Meyer, Techi, Domingues, Perez,
  Schmiedl, Lima, Yaguiu, Domingos, Busnello, Cardoso, Santos, Pimentel,
  Meneghetti, Tonidandel, and Junior]{robofei2018}
Marina~Y. Gonbata, et~al.
\newblock Robofei team description paper for robocup@home 2018.
\newblock \emph{RoboCup@Home 2018 Team Description Papers}, 2018.

\bibitem[Yaguchi et~al.(2017)Yaguchi, Tran, Takeda, Kochigami, Li, Sasabuchi,
  Furuta, Nagahama, Okada, and Inaba]{jsk2017}
Hiroaki Yaguchi, et~al.
\newblock Jsk@home: Team description paper for robocup@home 2017.
\newblock \emph{RoboCup @Home 2017 Team Description Papers}, 2017.

\bibitem[Martín et~al.(2018)Martín, Rodríguez, and
  Matellán]{gentlebots2018}
Francisco Martín, Francisco~J. Rodríguez, and Vicente Matellán.
\newblock Gentlebots 2018 team description paper.
\newblock \emph{RoboCup@Home 2018 Team Description Papers}, 2018.

\bibitem[Seib et~al.(2015)Seib, Manthe, Memmesheimer, Polster, and
  Paulus]{Seib2015}
Viktor Seib, et~al.
\newblock Team homer@unikoblenz—approaches and contributions to the
  robocup@home competition.
\newblock In \emph{Robot Soccer World Cup}, pages 83--94. Springer, 2015.

\bibitem[Hart et~al.(2017)Hart, Stone, Thomaz, and Niekum]{utaustinvilla2017}
Justin~W. Hart, et~al.
\newblock Ut austin villa robocup@home domestic standard platform league team
  description paper.
\newblock \emph{RoboCup @Home 2017 Team Description Papers}, 2017.

\bibitem[Lee et~al.(2017)Lee, Choi, Lee, Park, Choi, Baek, and
  Zhang]{aupair2017}
Beom-Jin Lee, et~al.
\newblock 2017 aupair team description paper.
\newblock \emph{RoboCup @Home 2017 Team Description Papers}, 2017.

\bibitem[Stanford(2011)]{stanford2011corenlponline}
Stanford.
\newblock Corenlp, 2011.
\newblock URL \url{http://nlp.stanford.edu:8080/corenlp/}.

\bibitem[Bastianelli et~al.(2016)Bastianelli, Croce, Vanzo, Basili, and
  Nardi]{Bastianelli2016}
Emanuele Bastianelli, et~al.
\newblock A discriminative approach to grounded spoken language understanding
  in interactive robotics.
\newblock In \emph{IJCAI}, pages 2747--2753, 2016.

\bibitem[Okada et~al.(2017)Okada, Yokoyama, Ogata, Inamura, Iwahashi, and
  Sugiura]{erasers2017}
H.~Okada, et~al.
\newblock Team er@sers[dspl] (toyota hsr) 2017 team description paper.
\newblock \emph{RoboCup @Home 2017 Team Description Papers}, 2017.

\bibitem[Kim et~al.(2018)Kim, Cauli, Vicente, Damas, Cavallo, and
  Santos-Victor]{Kim2018}
Jaeseok Kim, et~al.
\newblock ``icub, clean the table!'' a robot learning from demonstration
  approach using deep neural networks.
\newblock In \emph{2018 IEEE International Conference on Autonomous Robot
  Systems and Competitions (ICARSC)}, pages 3--9. IEEE, 2018.

\bibitem[Oliva et~al.(2004)Oliva, Wolfe, and Arsenio]{Oliva2004}
Aude Oliva, Jeremy~M Wolfe, and Helga~C Arsenio.
\newblock Panoramic search: the interaction of memory and vision in search
  through a familiar scene.
\newblock \emph{Journal of Experimental Psychology: Human Perception and
  Performance}, 30\penalty0 (6):\penalty0 1132, 2004.

\bibitem[Oliva and Torralba(2007)]{Oliva2007}
Aude Oliva and Antonio Torralba.
\newblock The role of context in object recognition.
\newblock \emph{Trends in cognitive sciences}, 11\penalty0 (12):\penalty0
  520--527, 2007.

\end{thebibliography}
